\documentclass[twocolumn,letterpaper]{IEEEAerospaceCLS}  

\usepackage{amsmath}
\usepackage[]{graphicx}    
\newcommand{\ignore}[1]{}  
\usepackage{siunitx}
\usepackage{amsfonts}
\newcommand{\fig}[1]{Figure~\ref{#1}}

\newcommand{\eq}[1]{(\ref{#1})}

\usepackage{epsfig} 
\usepackage{amssymb}  
\usepackage{subcaption}
\usepackage{caption}
\usepackage{color}
\usepackage{verbatim}
\usepackage[table,xcdraw]{xcolor}
\usepackage{xcolor}
\usepackage{multirow}
\usepackage{booktabs,caption}
\usepackage[noadjust]{cite}
\usepackage[flushleft]{threeparttable}
\usepackage[table,xcdraw]{xcolor}

\usepackage[normalem]{ulem} 

\newcommand{\RNum}[1]{\uppercase\expandafter{\romannumeral #1\relax}}
\makeatletter
\newlength\tmp@\newlength\t@mp
\newcommand{\comp}[3]
  {\mathop{ \settowidth\tmp@{$\displaystyle\mathop{#1}^{#3}_{#2}$}
  \hbox to \tmp@{\hss \settowidth\t@mp{$\displaystyle #1$}\setlength\t@mp{.45\t@mp}
  $\displaystyle\mathop{#1}^{\hspace\t@mp #3}_{\hspace{-\t@mp}#2}$
  \hss} }}

\usepackage{xspace}

\newcommand{\name}{SPLITTER\xspace}
\newcommand{\hname}{Hemi-SPLITTER\xspace}

\newcommand{\launch}{\textbf{Launch Phase}\xspace}
\newcommand{\flight}{\textbf{In-Flight Phase}\xspace}
\newcommand{\jump}{\textbf{Jump Phase}\xspace}
\newcommand{\landing}{\textbf{Landing Phase}\xspace}

\begin{document}
\title{Tethered Variable Inertial Attitude Control Mechanisms through a Modular Jumping Limbed Robot}

\author{%
Yusuke Tanaka,  Alvin Zhu, Dennis Hong\\
University of California, Los Angeles\\
Los Angeles, CA 90095\\
yusuketanaka@g.ucla.edu, alvinzhu2022@gmail.com, dennishong@g.ucla.edu
\thanks{\footnotesize 979-8-3503-5597-0/25/$\$31.00$ \copyright2025 IEEE}              
}

\maketitle

\thispagestyle{plain}
\pagestyle{plain}

\begin{abstract}
This paper presents the concept of a tethered variable inertial attitude control mechanism for a modular jumping-limbed robot designed for planetary exploration in low-gravity environments. The system, named \name, comprises two sub-$10$ kg quadrupedal robots connected by a tether, capable of executing successive jumping gaits and stabilizing in-flight using inertial morphing technology. Through model predictive control (MPC), attitude control was demonstrated by adjusting the limbs and tether length to modulate the system's principal moments of inertia. Our results indicate that this control strategy allows the robot to stabilize during flight phases without needing traditional flywheel-based systems or relying on aerodynamics, making the approach mass-efficient and ideal for small-scale planetary robots—successive jump. The paper outlines the dynamics, MPC formulation for inertial morphing, actuator requirements, and simulation results, illustrating the potential of agile exploration for small-scale rovers in low-gravity environments like the Moon or asteroids.
\end{abstract}

\tableofcontents

\section{Introduction}

In recent decades, planetary exploration has achieved significant milestones by deploying robotic rovers to physically explore and interact with the environments \cite{2020mars}. For instance, the successful operation of drones on Mars has enabled rapid observation over extensive areas, considerably enhancing mission efficiency and scientific output \cite{Ingenuity}. However, replicating such aerial exploration techniques on the Moon or more minor asteroids is not feasible due to the lack of a substantial atmosphere, requiring other means of flight control \cite{cat_like_jump}.

Limbed robots offer versatility in traversing diverse terrains, and their multi-modal movement capabilities enable them to perform walking \cite{spacebok}, climbing \cite{hubrobo}, \cite{yusuke_scaler_2022}, and leaping motions, \cite{cat_like_jump},  \cite{quad_jump_mpc}, making them suitable for missions in environments that would be challenging to traverse with conventional wheeled rovers. Additionally, the limbs can perform manipulation tasks, enhancing scientific research abilities. Explorations utilizing smaller but multi-agent systems have been investigated since they can more efficiently navigate low-atmosphere planets, and the implication of losing one robot is minor compared to traditional $100$ kg scale rovers \cite{winch_mini_rover} \cite{santra_risk_aware}. 
Under low-gravity conditions, jumping locomotion emerges as an efficient means of traversal \cite{spacebok}, particularly for smaller robots to quickly cover the larger area \cite{small_jump}. However, the more dynamic and agile jumping locomotion induces difficulty in attitude control during the flight phase, which is essential to land safely after the jumps \cite{quad_jump_mpc}, \cite{cat_like_jump}, \cite{spacebok}. 

\begin{figure}[t!]
\centering
\includegraphics[width=0.99\linewidth]{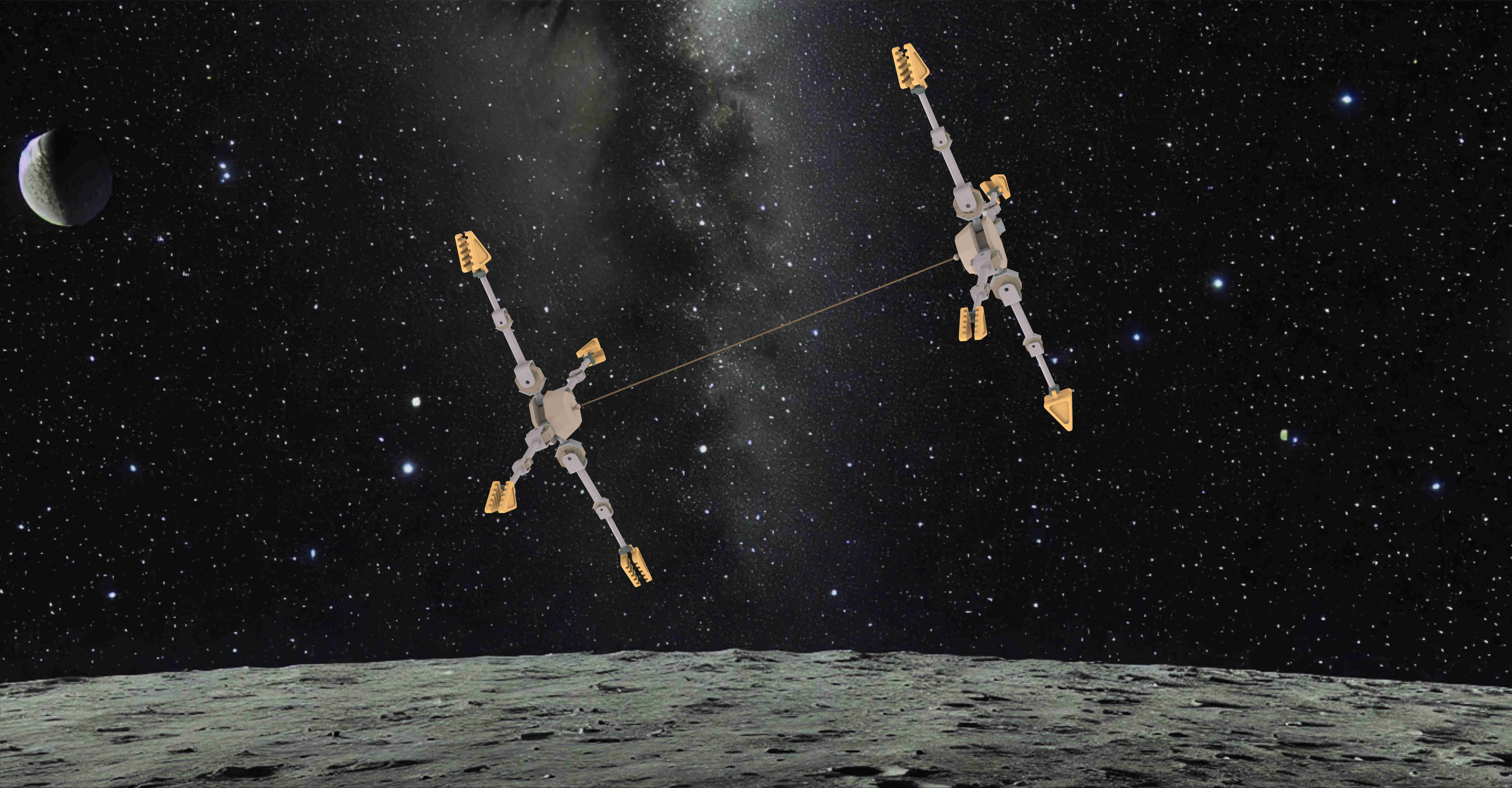}
\caption{A render of \name when arms spread out and tether elongated for inertial morphing attitude control.}
\label{flying1}
\end{figure}

Attitude control in satellites has been implemented through reaction wheels \cite{floating_space_capture}, which is not necessary or mass-efficient for limbed robotics, which has demonstrated attitude control through high inertia leg end-effector motions \cite{cat_like_jump}. Inertial morphing \cite{inertial_morphing}, where attitude control is done through variable inertia mechanisms, can achieve de-tumbling from more acrobatic 3D spins and unique rapid maneuvers through Garriott’s-Dzhanibekov’s Effect, or so-called tennis racket theorem \cite{inertia_morphing_attitude}. 

This paper investigates the feasibility of more successive jumping flight gait mechanisms using two sub-$10$ kg scale quadruped limbed robots that stabilize the system through model predictive controller-based inertial morphing attitude control during the flight phase. Two robots are connected via a tether, representing a dumbbell-like system, and the overall 3-degrees of freedom principle inertia control capability will demonstrate the gait's feasibility in terms of attitude control and limb actuator requirements. 

In summary, this paper presents the following contributions:

\begin{enumerate} 
\item \textbf{Inertial Morphing for Attitude Control with Model Predictive Control} We develop an inertial morphing technique, facilitated by limb adjustments, to achieve attitude control during flight phases without dedicated attitude control hardware such as a flywheel. 
\item \textbf{Tethered Dual-Robot System:} We introduce a conceptual sub-$10$ kg limbed robot for multi-modal locomotion and manipulation, capable of low-gravity planetary jumping flight gait.
\end{enumerate}

Our findings suggest that integrating successive jumping locomotion with inertial morphing control presents a viable and effective method for planetary exploration in low-gravity environments. This approach holds significant potential for rapidly exploring the Moon and asteroids with small-scale space robots without relying on aerodynamics.

\section{Related Works}
\subsection{Robotics in Space Planetary Exploration}
Large-scale rovers with rich scientific instruments have been successfully deployed for planetary exploration, such as Mars Perseverance\cite{2020mars}. Smaller-scale systems have seen the potential in space exploration and development, such as CubeSat and centimeter scale ChipSat \cite{manchester2013kicksat}. Thanks to increased interest from the private sector in space exploration, small-scale rovers have been gaining more attention \cite{santra_risk_aware}. For planetary exploration, various mini-rovers have been proposed to be specialized in tasks \cite{mini_rover_multi} and more efficiently observe large areas \cite{winch_mini_rover}. Ingenuity \cite{Ingenuity} is a notable example of successfully exploring a wider area on Mars than any previous rovers by flying over the surface. Meanwhile, a fixed type of modality has seen limitations, and more adaptable mechanisms are being explored, such as a skid-steer with a drone \cite{fly_mars_sample}, and an eel-like robot that can move, stand, and climb \cite{vaquero2024eels}. 
The \name concept leverages the adaptability of quadruped-limbed robots, which can use two limbs for manipulation tasks while maintaining stability by standing on the remaining limbs \cite{alphred}.

\subsection{Limbed Robotics in Space}
Limbed robots are prominent systems that can be versatile and enable exploration on previously impossible surfaces through their locomotion and grasping capabilities, such as HubRobo \cite{hubrobo}, LEMUR3 \cite{lemur3}, and SCLAER  \cite{yusuke_scaler_2022}. Relatively larger-scale-limbed robot RoboSimian proposed to conduct climbing with a SpineyHand \cite{spiny_hand} or wheel-on-limb locomotion \cite{wheel_on_limb} to enhance mobility on rocky and icy terrains.
The multi-modal nature of the limbed robots can traverse rough trains, overcome obstacles, and leap to move through low-gravity environments effectively \cite{spacebok}. We investigate the jumping and flying potential of the limbed robot system, which stabilizes the flight phase through inertial morphing by changing the limb configuration and tether length.

\subsection{Modular and Tethered Robotics in Space}
Modularity and reconfigurability are other ways of increasing adaptability by morphing the mechanical design itself \cite{makabe_mophable_modular}. A tethered dual two-wheeled rover, DuAxel \cite{duaxel}, has been designed to reduce the complexity while achieving multi-modality and a wide range of task applications, including 4-wheel driving, 2-wheel driving, tethered climbing, sample analysis, and lunar radio construction \cite{moon_Crater_radio}. A quadruped robot with tethered grippers illustrated an interesting approach to enhance the robot's stride and reach for discrete, rough environment locomotion \cite{tetherbot}. 
Tethered multi-agent robots can address limitations that arise from the constrained resources of mini-rovers through hyper-tethering \cite{hyper_tether}, or by working collaboratively through winch tether mechanisms \cite{winch_mini_rover}, \cite{follower_tether_rover}. The \name concept utilizes two quadrupeds connected by a tether, which can adjust the inertia of the entire system and provide additional benefits as described above.

\subsection{Attitude Control}
The attitude control in 3D space is a well-researched area, as shown with systems such as satellites \cite{floating_space_capture} and drones \cite{lopez_quat}, \cite{quat_mpc}. MPCs have been applied to limbed robots \cite{tanaka2023scaler}, which considers the nonlinear dynamics of 3D systems with constraints. Reinforcement learning has embraced the use of the high-inertia end-effector limb to realize a safe jump and landing \cite{cat_like_jump}. Inertial morphing \cite{inertial_morphing} extends the variable inertia concept to conduct more dynamic attitude control \cite{inertia_morphing_attitude}. In this paper, we investigate if the inertial morphing using and MPC can help stabilize \name's jumping flight phase, which can change the inertia by adjusting the limb end-effector and tether length.

\section{\name System Dynamics}
\begin{figure}
\centering
\includegraphics[width=0.99\linewidth]{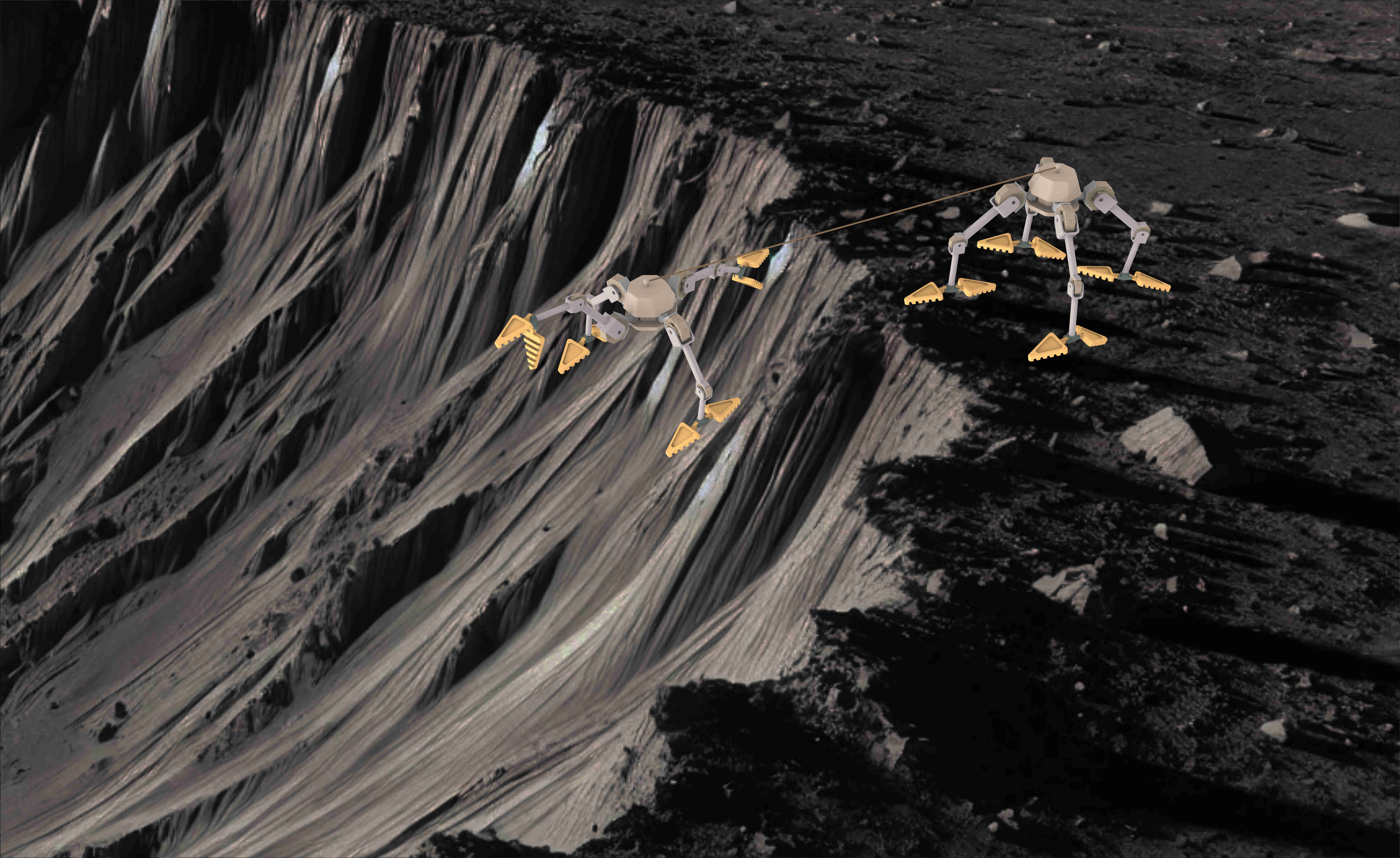}
\caption{\name One quadruped in biped mode with two manipulators inside of a creator and the other quadruped as an anchor.}
\label{bipid_manip}
\end{figure}

\name is the proposed tethered limbed robot mechanisms that consist of two sub-robots called \hname connected via a single tether. Each sub-robot in \name is designed as an independent system capable of performing tasks autonomously, including locomotion and manipulation. However, when connected via the tether, they can collaborate, combining their capabilities to tackle more complex tasks, such as cooperative lifting, enhanced mobility, or traversing challenging environments.
Our unique two sub-robot quadrupedal design allows us to perform tasks such as bipedal locomotion with simultaneous manipulation, with the other sub-robot acting as the anchor, as shown in \fig{bipid_manip}. \hname is intended to be equipped with spine grippers as seen in \cite{hubrobo}, \cite{spiny_hand}, \cite{GOAT}, which are adequate for rough rocky environments. 

\begin{figure}
\centering
\includegraphics[width=0.99\linewidth]{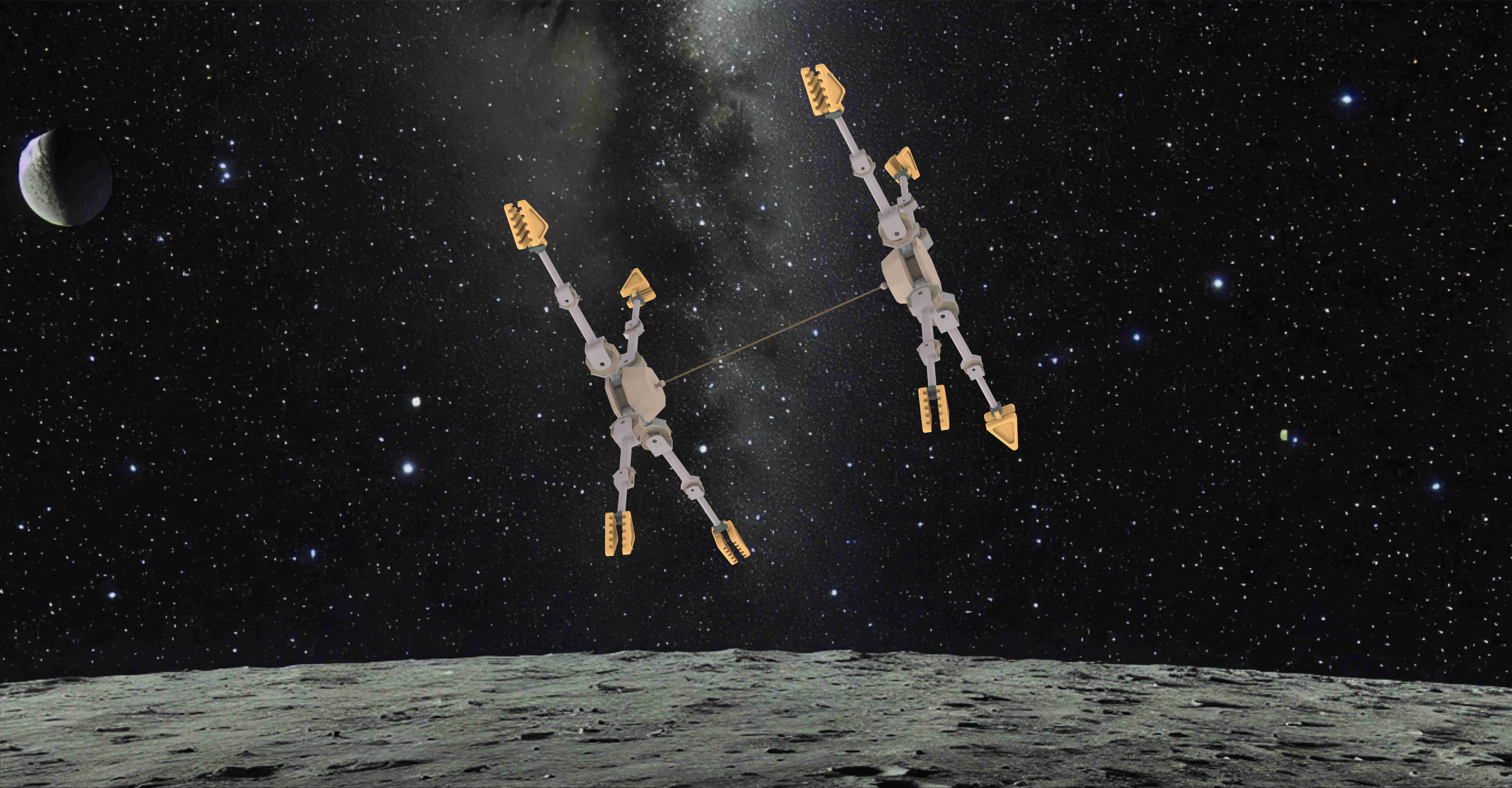}
\caption{\name Arms spread out and tether shortened.}
\label{flying2}
\end{figure}

\begin{figure}
\centering
\includegraphics[width=0.99\linewidth]{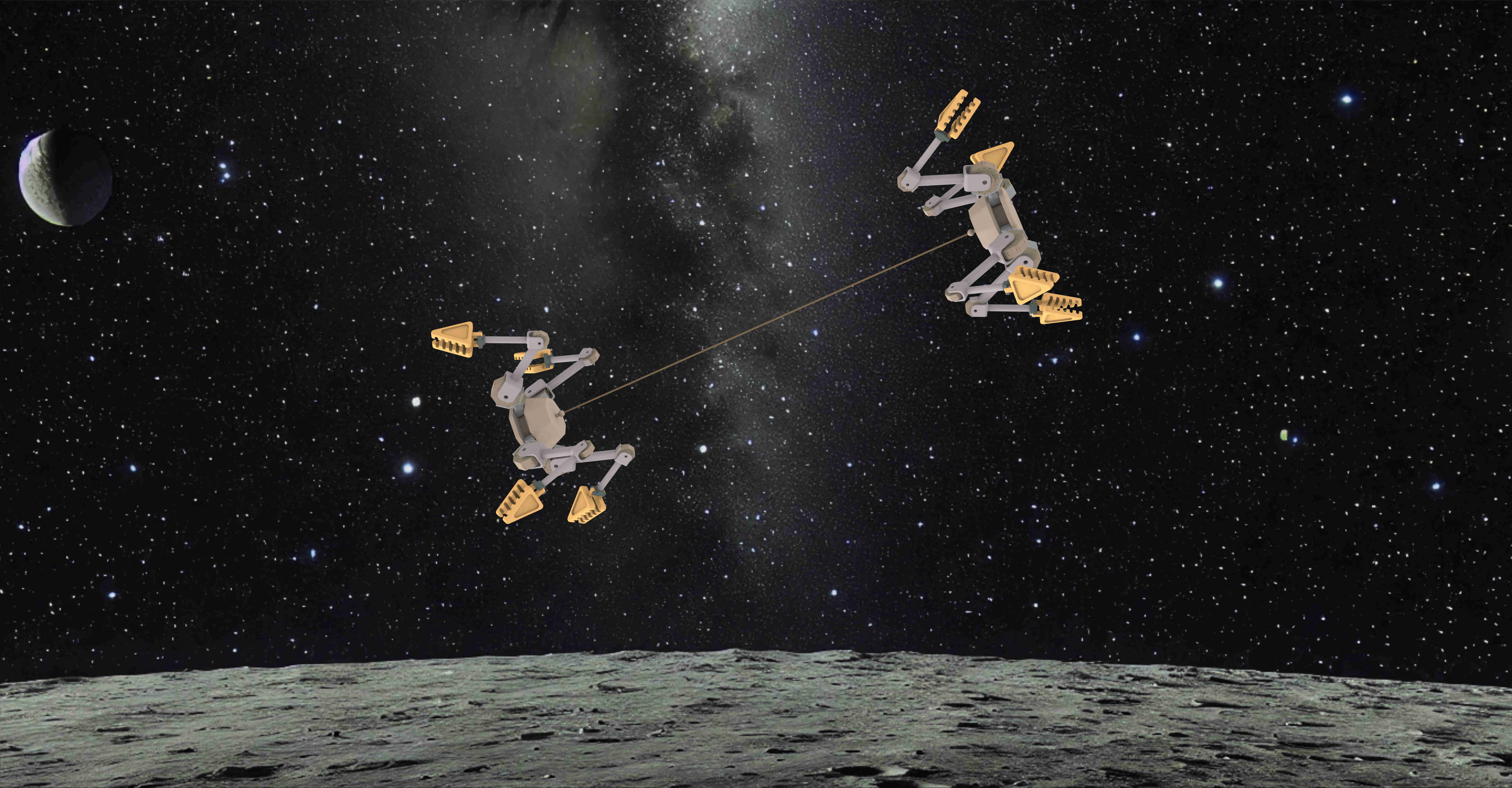}
\caption{\name Arms shortened and tether elongated.}
\label{flying3}
\end{figure}

\begin{figure}[t!]
\centering
\includegraphics[width=0.99\linewidth]{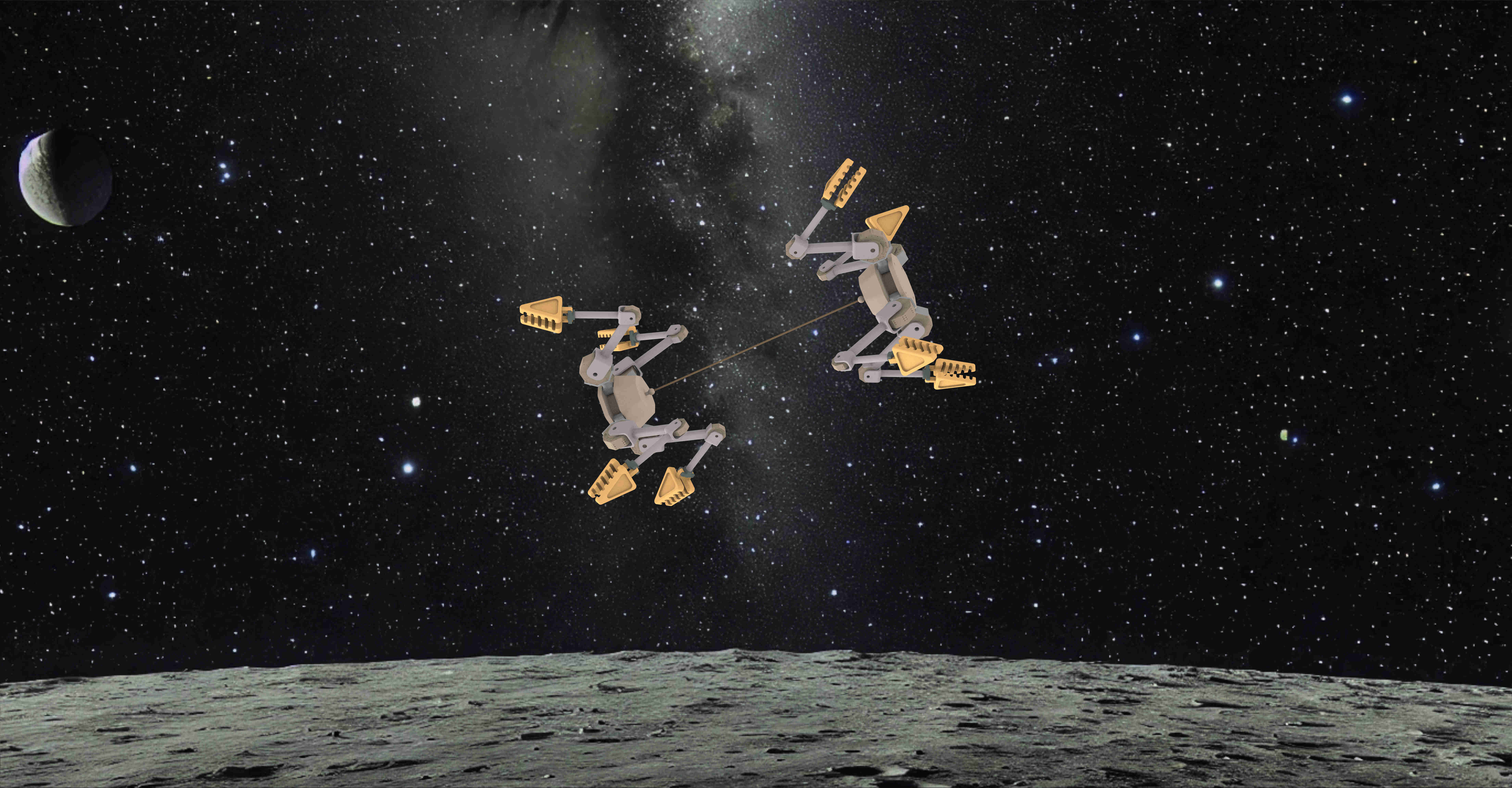}
\caption{\name Arms shortened and tether shortened.}
\label{flying4}
\end{figure}

Additionally, this design allows us to conduct attitude control through inertial morphing, where \name changes the principle axis inertia by changing the tether length and reorienting its joints. By moving the joints with high-inertia end-effectors at the extremities, \name can effectively shift its mass distribution, enabling precise control over its inertia for complex maneuvers in space as shown in Figure \ref{flying1}, \ref{flying2}, \ref{flying3}, \ref{flying4}. 

This section discusses mechanical requirements and actuator selections that enable highly dynamic jumping flight locomotion, inertial morphing, ground exploration, and quasi-static tethered climbing.

\subsection{Jumping Flight Gait Mechanisms}
Our proposed gait mechanism includes the following phases. 1) \launch, 2) \flight, 3) \jump, 4) \landing

In \launch, each side of \hname jumps while adding a rotational motion, whose energy will be later used for inertial morphing. 
In \flight, \name controls the entire system inertia by adjusting the tether and the eight-limb end-effector distances with respect to the base of \hname. This will allow \name to stabilize and control the angular velocities in flight. 
In \jump, one of \hname makes contact with the ground and jumps, providing a thrust force to the system to continue the flight and increasing the angular momentum. 
\name repeats the \flight and \jump to continue the journey till it needs to land and rest. 
Towards \landing, \name de-thrusts the system, and the attitude controller slows down the angular velocity for landing.  

\subsection{Variable Inertial System}
Our approach models \name as a variable inertial system that can actively adjust its principal moments of inertia by changing the configuration of its limbs and the length of the tether connecting the two sub-robots. 

\begin{figure}[t!]
\centering
\includegraphics[width=0.7\linewidth]{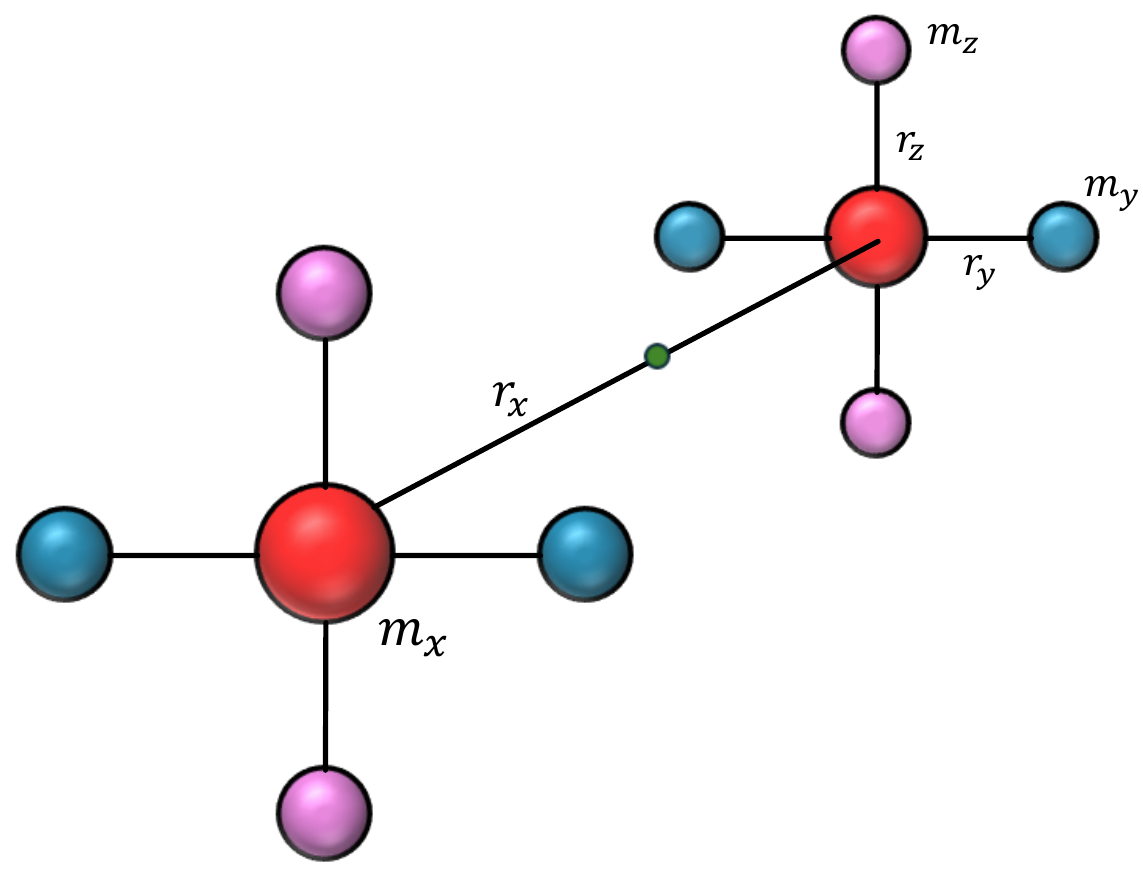}
\caption{\name simplified 5-dumbbell-like system representation.}
\label{dumbbell_5}
\end{figure}

The \name inertial model is assumed and simplified as a rigid-body five dumbbell-lumped masses, as shown in \fig{dumbbell_5}. A tethered spacecraft is conventionally assumed to be a rigid body dumbbell with point masses when the tether is taut with negligible elasticity \cite{sanyal2005stability}. 
Both \hname does not necessarily need to be in a symmetric configuration such as shown in \fig{flying3}, but the effective inertia needs to align with the model shown in \fig{dumbbell_5}. For this reason, the proposed inertial morphing MPC controls the principle axis inertia. Then, the lower-level inverse kinematics controller solves the robot joint poses. 

Where the tether's axial and torsional stiffness are reasonably high, then it can be assumed to be rigid. Hyper tether \cite{hyper_tether} or significantly stiff tether in relatively short length can hold this assumption. Otherwise, another actuation controlling the twisting may be necessary.
More discussions on the tether slack condition are in Section \ref{sec:slack} to avoid the system being compressed. 

Since the system is symmetric in all planes, we assume the non-diagonal element of the inertial matrix is zero.


The moment of inertia matrix is represented in Eq. \ref{eq:inertia}.

\begin{equation}
\label{eq:inertia}
I =     \left[\begin{array}{ccc}
I_{x x} & 0 & 0 \\
0 & I_{y y} & 0 \\
0 & 0 & I_{z z}
\end{array}\right]
\end{equation}
$I_{x x}, I_{y y}$ and $I_{z z}$ are the principle axis moments of inertia about the $x-, y$-, and $z$-axes, respectively. 
The principal moments of inertia are derived as in Eq. \ref{eq:principal}.

\begin{align}
\begin{split}\label{eq:principal}
I_{x x} &= 4 m_y r_y^2 + 4 m_z r_z^2 \\
I_{y y}&=\left(2 m_x+4 m_y+4 m_z\right) r_x^2+4 m_z r_z^2 \\
I_{z z}&=\left(2 m_x+4 m_y+4 m_z\right) r_x^2+4 m_y r_y^2
\end{split}
\end{align}
$m_x, m_y, m_z$ are the lumped masses of the components along the $x$-axis, $y$-axis, and $z$-axis.
$r_x, r_y, r_z$ are the radii representing the distances from the system's center of mass along the principal axes.

Solving for the radii of the principal axis given the moment of inertia:
\begin{align}
r_x&= \frac{1}{2} \sqrt{\frac{I_{zz} - I_{xx} + I_{yy}}{m_x + 2m_y + 2m_z}}\\
r_y&=\sqrt{\frac{I_{x x}-I_{y y}+I_{z z}}{8 m_y}}\\
r_z&=\sqrt{\frac{I_{y y}-I_{z z}+I_{x x}}{8 m_z}}
\end{align}

\subsubsection{Tension}
The tension in the tether can be written as,
\begin{equation}
T=m_x \omega_y^2 r_x+m_x g \sin \theta
\end{equation}
$\omega_y$ is the angular velocity around $y$, and $\theta$ represents the dumbbell angle in 2D. 

\subsection{Actuator Selection Considerations}
\label{actuator_considerations}

The selection of actuators for limbed robots involves evaluating both torque and speed requirements, as well as considering the robot's dynamics and desired performance characteristics. We utilize a simplified calculation of the model to help us choose a suitable actuator.

\subsubsection{Torque and Angular Velocity Requirements\label{sec:slack}}

For each of the sub-robots to achieve lift-off and reach a height of $5$ m, given each sub-robot is $\leq10$ kg and limbs are $0.4$ m long, the actuators must overcome the Moon's gravitational force and accelerate enough to reach a certain lift-off velocity. The force is distributed across four legs, and the torque is mainly outputted by the larger hip actuator, thus we can simplify the system by assuming a four joint system. The force required for each leg, \( F_{\text{leg}} \), and the torque required by the hip joint, \(T_{\text{hip}} \), is determined by the robot's mass \( m \), gravitational acceleration \( g_{moon} \), and the acceleration required to reach the desired takeoff velocity \( v_0 \). The resulting takeoff velocity and acceleration are $4.0\text{ m/s}$ and $21.7\text{ m/s}^2$, respectively. This results in a force per leg being $52.5$ N, and torque for each hip joint being $10.5$ Nm. The maximum angular velocity required for each hip joint results in $270.3$ RPM.

\section{Dynamics Simulation}
We implement our custom rigid body dynamics simulator with contact modeling since the existing simulators have exhibited issues in conserving energy after changing the rigid body geometries and tether lengths. 
The simulation considers the following models to calculate the rotational and translational dynamics of the 5-dumbbell-like system \fig{dumbbell}.

\subsubsection{Euler's Equations of Motion:}
The dynamics simulation is driven by Euler's rigid body dynamics.  
Euler's equations govern the rotational dynamics of a rigid body, shown in Eq. \ref{eq:moment} and \ref{eq:linear}.

\begin{equation}\label{eq:moment}
    \mathbf{I} \dot{\boldsymbol{\omega}} + \boldsymbol{\omega} \times (\mathbf{I} \boldsymbol{\omega}) = \mathcal{M}
\end{equation}

\begin{equation}\label{eq:linear}
M \dot{\mathbf{v}} = \mathbf{F} - B \mathbf{v} - M \mathbf{g}
\end{equation}
Where $\omega$ and $\dot{\omega}$ are angular velocity vectors and its derivative. $\mathcal{M}$ is the angular momentum. 
$M$ is the diagonal mass matrix, $V$ and $\dot{v}$ are the velocity vector and its derivative. $F$ is the total external applied force and $B$ is a damping coefficient.

To obtain the system's 3D motion, we use the 4th-order Runge-Kutta method (RK4). RK4 is an explicit method used to solve ordinary differential equations, including the dynamics of the Dumbbell-like system.
Each RK4 coefficient is an intermediate value obtained from the derivatives at different points in the timestep. RK4 is relatively stable and allows the simulation timestep to be larger than the implicit Euler methods.

\subsubsection{Contact Dynamics}
In our simulation, the contact dynamics play a significant role in capturing the successive jumping and flight phases, which introduces hybrid dynamics. We employ a spring-damper contact model \cite{contact_optimization_legged}. 

\begin{equation}
F_c = s \cdot \delta_x+c \cdot \dot{\delta_x}
\end{equation}

where $F_c$ is the contact force, $\delta_x$ is the penetration distance, $s$ and $c$ are the ground stiffness and damping, respectively. Consequently, this method allows a rigid object to violate the ground constraint or small penetration. More naive approaches, such as forcing the sticking or ground height, can violate energy conservation throughout the jumping sequence.

The frictional force is subject to the friction cone as:
\begin{equation}
    \sqrt{F_{f_x}^2+F_{f_y}^2} \leq \mu \cdot F_{n}
\end{equation}
Where $F_{f}$ is the frictional force in $x$ and $y$ the ground plane, $\mu$ is the Coulomb friction coefficient, and $F_n$ is the normal force due to contact.


\section{Model Predictive Control}

\begin{figure}
\centering
\includegraphics[width=0.99\linewidth]{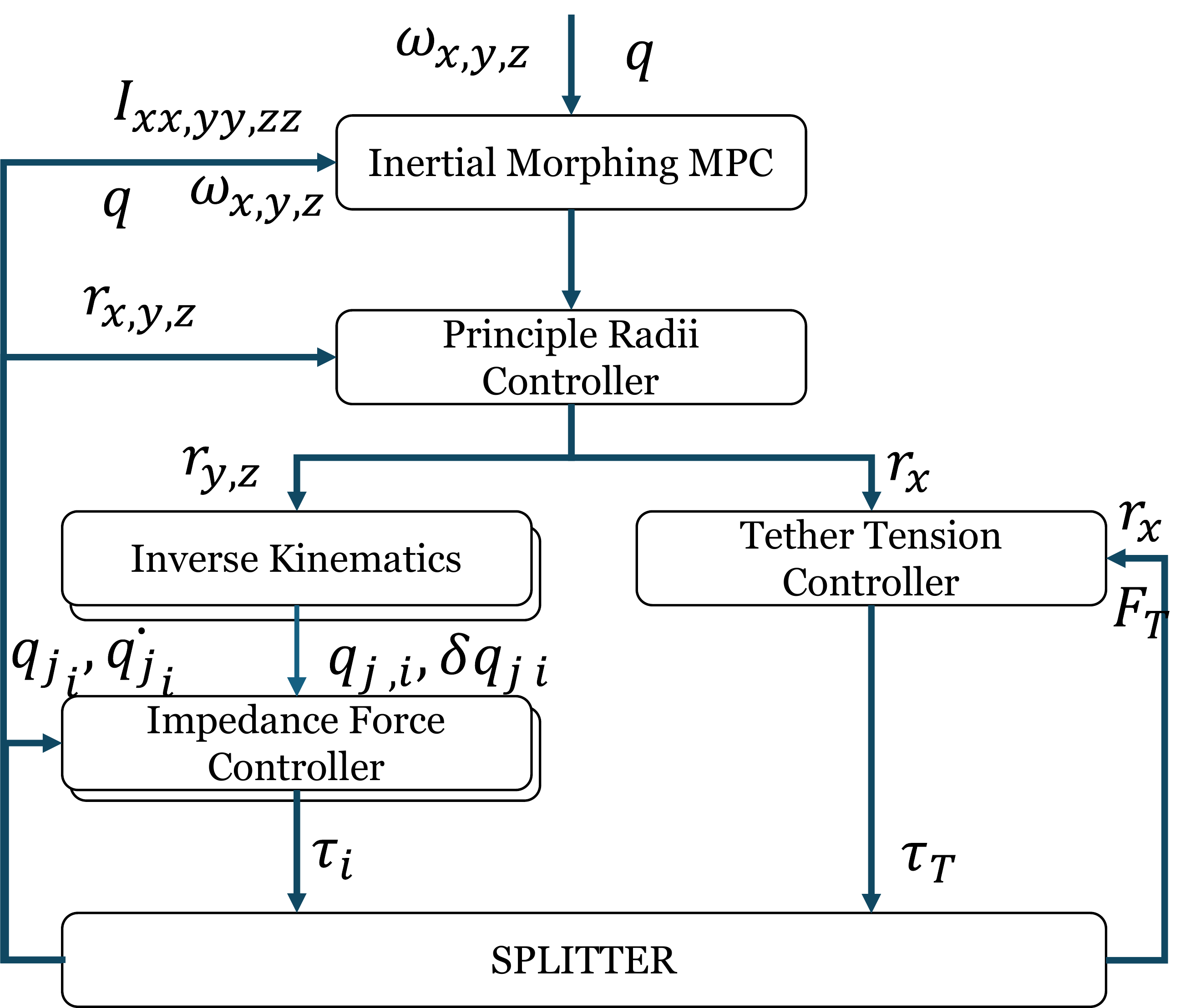}
\caption{\name jumping flight control framework structures.}
\label{controller}
\end{figure}
This section covers the mathematical formulation of our inertial morphing MPC. The inertial morphing MPC is a part of the overall control framework as shown in \fig{controller}. In this paper, inverse kinematics and impedance controllers are not covered, but they are well-researched in limbed robotics \cite{alex_admittance}, \cite{uno_admittance}. 

\subsubsection{Variables}
Since \name is a variable inertial system with 3D orientations, our state variables are defined as:  
\begin{equation}
x_k = \begin{bmatrix} \omega_k & q_k & I_{i, k} \end{bmatrix}^T \quad \forall i=xx,yy,zz
\end{equation}

Where $\omega_k \in \mathbb{R}$$^3$  represents the angular velocity vector at time step \(k\),$q_k \in \mathbb{R}$$^4$ is the quaternion representing the orientation at time step \(k\), and $I_{i,k} \in \mathbb{R}$$^3$ is the diagonal elements of inertia matrix representing the principle axis inertia at time step \(k\).

\subsubsection{Control Variables:}
We choose to use the principle axis inertia directly as control variables since \name is symmetric and the inertia matrix is diagonal.
\begin{equation}
u_k = \Delta I_{i,k}
\end{equation}
where $\Delta I_{i,k} \in \mathbb{R}$$^3$ represents the change in the moments of inertia in the principle axis at time step \(k\).  Later, we discuss how we bound the control variables.

\begin{figure*}[t!]
\centering
\includegraphics[width=0.9\linewidth]{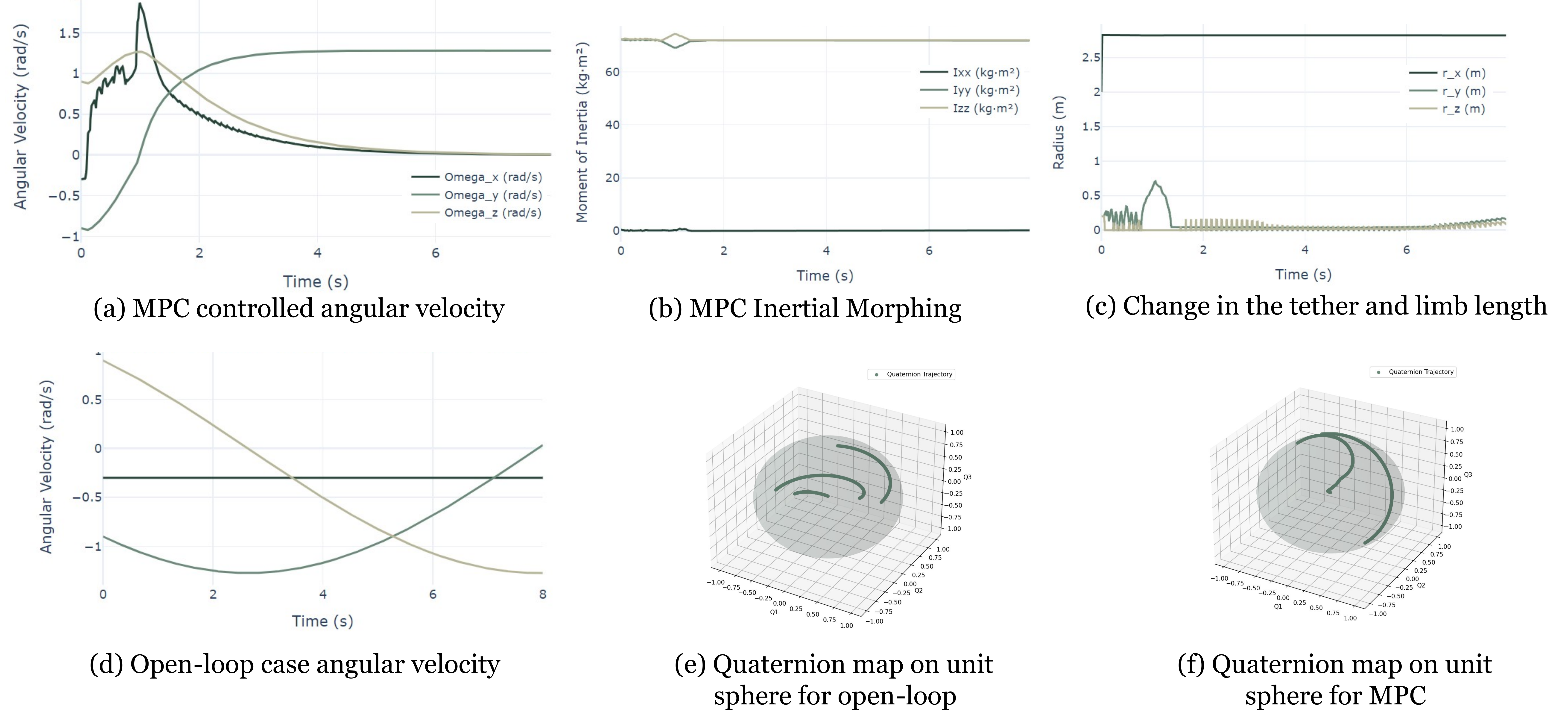}
\caption{Comparisons of inertial morphing MPC and uncontrolled cases. }
\label{omega}
\end{figure*}

\subsection{Discrete-Time Dynamics}
\subsubsection{Angular Momentum Dynamics}
Dynamics due to the angular momentum is given as:
\begin{align}
I_{k+1} \omega_{k+1}&=I_k \omega_k+\Delta t \cdot \tau_k\\
\tau_k&=\omega_k \times\left(I_k \omega_k\right)
\end{align}
where $I_k \in \mathbb{R}$$^3$ is the diagonal matrix representing the moments of inertia at time step \(k\), \( \Delta t \) is the time step, and $\tau_k$ is the torque due to the rotation.

Since inertial morphing consists of highly dynamic and acrobatic 3D rotation, we take quaternion integration. 
\begin{equation}
q_{k+1} = \frac{q_k + \dfrac{1}{2} q_k \otimes \omega_{\text{quat},k} \Delta t}{\left\| q_k + \dfrac{1}{2} q_k \otimes \omega_{\text{quat},k} \Delta t \right\|}
\end{equation}
where $q_k \in \mathbb{R}$$^4$ is the quaternion at time step \(k\), \( \omega_{\text{quat},k} = \begin{bmatrix} 0,  \omega_k \end{bmatrix}^T \) is the quaternion form of the angular velocity \( \omega_k \). Note that the resultant quaternion is normalized. 
The physics simulation uses the full angular momentum equation, but the MPC propagates the angular momentum using a linearized model and Euler integral. 

\subsubsection{Inertia Update} 
The principle axis inertia elements are updated as: 
\begin{equation}
I_{i, k+1} = I_{i,k} + \Delta I_k \forall i=xx,yy,zz
\end{equation}

\subsection{Cost Function}
The MPC minimization objectives are tracking, control effort, and terminal costs. Thus, the objective is denoted in \ref{eq:cost}. 

\begin{equation}\label{eq:cost}
\min_{\omega_k, q_k, \Delta I_k} ( J_{\omega} + J_{q} + J_{c} + J_{t} )
\end{equation}

$J_{\omega}$ and $J_{q}$ are the angular velocity and quaternion tracking costs described in Section \ref{sec:tracking_cost}, $J_{c}$ is the control effort cost in Section \ref{sec:control_cost}, and $J_{t}$ is the terminal state cost in Section \ref{sec:terminal_cost}. Tracking costs encourage the MPC to follow the reference trajectory, the control effort cost to minimize the changes in the control inputs, and the terminal cost encourages to reach the goal state at the end of the prediction time horizon. All costs are weighted.

\subsubsection{Tracking Cost\label{sec:tracking_cost}}
The angular velocity tracking cost over time horizon $N$ is denoted as: 
\begin{equation}
J_{\omega} = \sum_{k=0}^{N-1} \left( \omega_k - \omega_{\text{ref},k} \right)^T Q_{\omega} \left( \omega_k - \omega_{\text{ref},k} \right)
\end{equation}
where $\omega_{\text{ref},k} \in \mathbb{R}$$^3$ represents the reference angular velocity at time step \(k\), and $Q_\omega \in \mathbb{R}$$^{3 \times 3}$ is the diagonal weight matrix for the angular velocity tracking cost. 

The 3D orientation tracking cost is:
\begin{equation}
J_{q} = \sum_{k=0}^{N-1} Q_{q} \left( 1 - \Re(e_{q, k}) \right)
\end{equation}
where \(e_{q,k} = q_{\text{ref},k}^* \otimes q_k\) is the quaternion error between the reference and the current quaternion, and \(\Re(e_{\text{q},k})\) is the scalar (real) part of the quaternion error. \( Q_{q} \) represents the weight for the orientation tracking cost.

\subsubsection{Control Effort Cost\label{sec:control_cost}}
To smooth out the control input to the system and reduce the control efforts, we minimize the change in the control as:
\begin{equation}
J_{c} = \sum_{k=0}^{N-1} \left( \Delta I_k \right)^T R \left( \Delta I_k \right)
\end{equation}
where $R \in \mathbb{R}$$^{3 \times 3}$ is the diagonal weight matrix for control effort.

\subsubsection{Terminal State Cost\label{sec:terminal_cost}}
The terminal cost encourages MPC to minimize the tracking error, particularly at the horizon's end. This terminal cost was necessary, and otherwise, the MPC solution converged to a constant offset from the reference states. 

\begin{equation}
\begin{split}
J_{t} = \left( \omega_N - \omega_{\text{ref},N} \right)^T Q_t \left( \omega_N - \omega_{\text{ref},N} \right) \\
+ Q_{t,q} \cdot \left( 1 - \Re(e_{q,N}) \right)
\end{split}
\end{equation}

Where $Q_t \in \mathbb{R}$$^{3 \times 3}$ is the terminal cost diagonal weight matrix for angular velocity and $Q_{t,q}$ is the weight corresponding to the quaternion terminal cost. 

\subsection{Constraints}
All decision variables are subject to their initial state constraints, which are obtained through state feedback. 
The change in principle axis inertia is constrained as 
\begin{equation}
- \Delta I_{i,\text{max}} \leq \Delta I_{i, k} \leq \Delta I_{i, \text{max}} \quad \forall i=xx,yy,zz
\end{equation}
where \( \Delta I_{\text{max}} \) is the maximum allowed change in inertia per time step. 
The control variable is bounded by:
\begin{equation}
0 < I_{i,k} \leq I_{i, max} \quad \forall i=xx,yy,zz
\end{equation}
where \( I_{i, \text{max}} \) represents the maximum physical value of the moments of inertia. 
Since the lower level control variables are $r_x$, $r_y$, and $r_z$, this bound is subject to \eq{eq:principal}. The Eq. \eq{eq:principal} can be considered a circle equation, and $I_i$ should be within the circle to strictly apply these bounds. 
Alternatively, the maximum $I_i$ is calculated with the maximum $r_x$, $r_y$, and $r_z$ using \eq{eq:principal} and applying saturation. 
Eq. \eq{eq:principal} makes the bound of $\Delta I_i$ nonlinear as well. 
Hence, a conservative smaller bound is applied on $\Delta I_i$ to avoid this nonlinearity but ensure MPC solutions do not exceed the bound loosely.

\section{Simulation and Results}
\subsection{Inertial Morphing MPC}
In this section, we explore the capability of inertial morphing MPC and its hyperparameters. 
Due to the nature of inertial morphing for attitude control, the system will have angular momentum, meaning we cannot set all angular velocity components to zero.

\subsubsection{Angular velocity tracking}

The parameters for the simulation and the MPC are shown in Table. \ref{tab:mpc_omega}. The weights for quaternion tracking were set to zero. Terminal cost weight for $\omega_y$ was set low to prioritize the other axes' convergence.

For around $x$ and $z$ axes, the inertial morphing MPC has successfully brought the system rotational to a steady state as shown in \fig{omega}a with residual errors of $0.004$ and $0.006$ rad/s, with the reference $0$ rad/sec. These axes are more critical compared to the $y$ axis rotation since those are unwanted motions in the jumping flight. Although there was no overshoot at the steady state, it does overshoot significantly in the process of inertial morphing. 

For around $y$ axis, the inertial morphing MPC time constant was $1.3$ s and converged to the steady state of $1.28$ rad/s. This is significantly overshooting since the reference tracking weight to this axis is lower. The settling time is $2.8$ s. At $t=0$ and $t=8$ s, the rotational energy was conserved.

The control efforts shown in \fig{omega}b,c confirm that the changes in $r_{x}$, $r_{y}$, $r_{z}$ are reasonably in the range of \name motion. Most control efforts are done by the $r_{x}$ and $r_{z}$, representing the \hname limb motions. The maximum G experienced at the \hname during the attitude control is $0.47$ G. 

The open loop case in \fig{omega}d,e shows the uncontrolled free spins. Although only angular velocity is regulated, the \fig{omega}f indicates the motion is regulated as seen in \fig{omega}a.

\subsubsection{3D orientation tracking}
Here, experiments inertial morphing MPC with active quaternion tracking. Table \ref{tab:orientation_tracking} shows the parameter used in this case.  

\begin{figure*}
\centering
\includegraphics[width=0.75\linewidth]{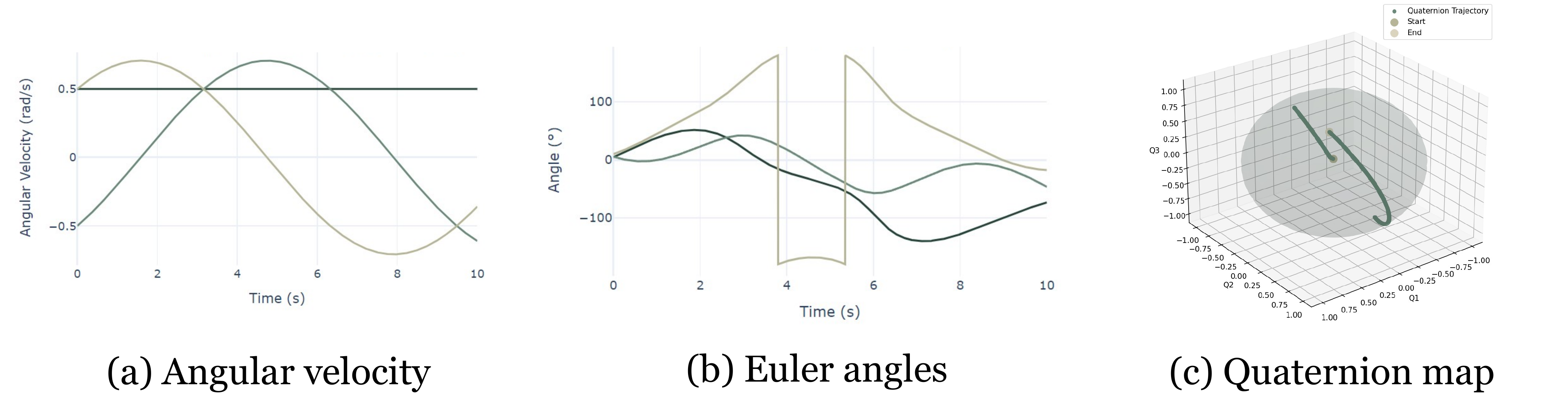}
\caption{The uncontrolled open-loop case.}
\label{quat_following_open}
\end{figure*}

\begin{figure*}
\centering
\includegraphics[width=0.9\linewidth]{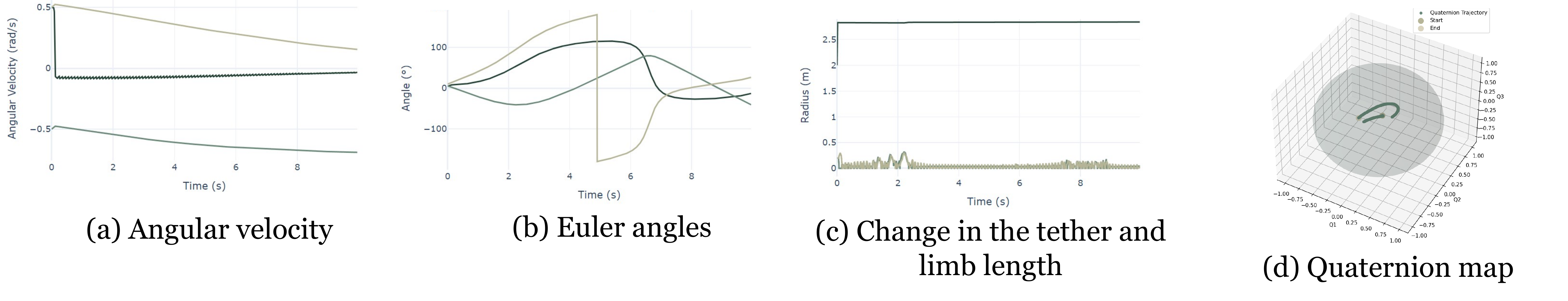}
\caption{The inertial morphing MPC without quaternion tracking.}
\label{quat_following_vel}
\end{figure*}

\begin{figure*}
\centering
\includegraphics[width=0.9\linewidth]{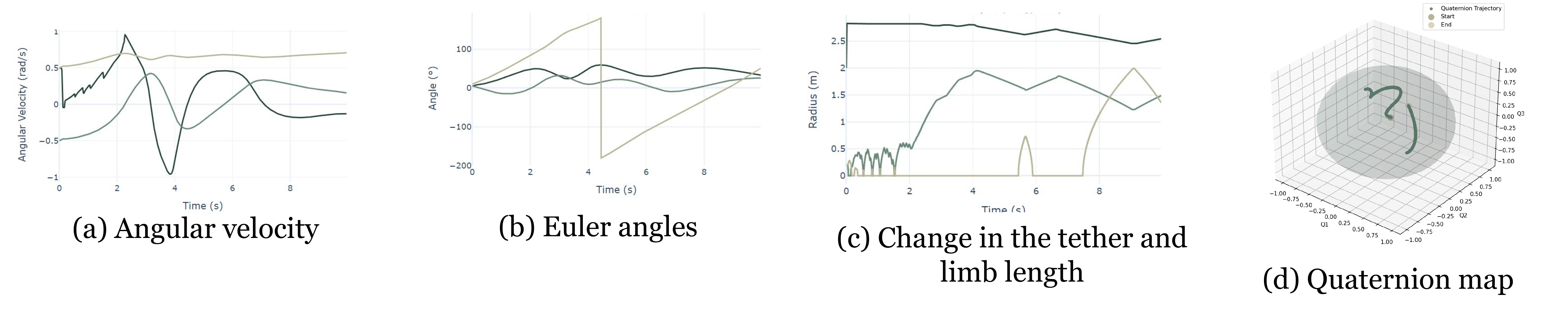}
\caption{The inertial morphing MPC with quaternion tracking.}
\label{quat_following}
\end{figure*}

\fig{quat_following_open}, \fig{quat_following_vel}, and \fig{quat_following} compare the system's dynamic motion without a controller and MPC with and without quaternion tracking. Quaternion tracking still allows \name to rotate around one axis, as with the previous case. The quaternion reference is nonconstant, which was calculated based on the angular velocity reference. Due to relatively smaller initial angular momentum, both MPC cases converged more slowly. \fig{quat_following_open}a, \fig{quat_following_vel}a, and \fig{quat_following}a show that the angular velocity tracking is better tracked without quaternion tracking. \fig{quat_following_open}b, \fig{quat_following_vel}b, and \fig{quat_following}b compares the \name Euler angles converted from quaternion and the corresponding quaternion map is shown in \fig{quat_following_open}c, \fig{quat_following_vel}d, and \fig{quat_following}d. \fig{quat_following_vel}c and \fig{quat_following}c are the calculated tether and limb lengths based on the inertial morphing MPC control inputs. The quaternion tracking requires a more aggressive and broader range of principle inertia. 
When tracking quaternions, the inertia must be adjusted significantly more than angular velocity tracking, indicating that \hname requires longer limbs if strict angular orientation control is necessary. We expect that the controlling angular velocity is sufficient for landing \name, since limbed robots have a large spherical workspace to land from the side, but this topic needs further research.

\begin{table*}[h!]
\centering
\caption{Simulation and MPC Parameters}
\label{tab:mpc_omega}
\begin{tabular}{|l|l|}
\hline
\textbf{Parameter} & \textbf{Value} \\ \hline
Simulation timestep ($\Delta t_{\mathrm{sim}}$) & 0.005 s \\ \hline
MPC solve frequency ($\Delta t_{\mathrm{MPC}}$) & 0.1 s \\ \hline
MPC prediction timestep & 0.01 s \\ \hline
Prediction horizon ($N$) & 50 steps \\ \hline
Masses & $m_x = 3.0$, $m_y = 1.5$, $m_z = 1.5$ kg \\ \hline
Initial radii & $r_x = 2.8$, $r_y = r_z = 0.2$ m \\ \hline
Damping coefficient ($b$) & 0.001 \\ \hline
Initial angular velocity ($\omega_0$) & $[-0.3, -0.9, 0.9]$ rad/sec \\ \hline
Reference angular velocity ($\omega_{ref}$) & $[0, 1, 0]$ rad/sec \\ \hline
Angular velocity tracking cost ($Q_\omega$) & diag([100, 100, 100]) \\ \hline
Control effort cost ($R$) & diag([0.01, 1, 1]) \\ \hline
Angular velocity terminal cost ($Q_t$) & diag([100, 10, 100]) \\ \hline
Quaternion tracking cost ($Q_{q}$) & 0 \\ \hline
Quaternion terminal cost ($Q_{t,q}$) & 0 \\ \hline

\end{tabular}
\end{table*}

\begin{table*}[h!]
\centering
\caption{3D Orientation Tracking Parameters}
\label{tab:orientation_tracking}
\begin{tabular}{|l|l|}
\hline
\textbf{Parameter} & \textbf{Value} \\ \hline
Initial angular velocity ($\omega_0$) & $[0.5, -0.5, 0.5]$ rad/sec \\ \hline
Reference angular velocity ($\omega_{ref}$) & $[0, 1, 0]$ rad/sec \\ \hline
Initial orientation & $[5, 5, 15]$ degrees (in quaternion) \\ \hline
Angular velocity tracking cost ($Q_\omega$) & diag([100, 10, 100]) \\ \hline
Control effort cost ($R$) & diag([0.001, 1, 1]) \\ \hline
Angular velocity terminal cost  ($Q_t$) & diag([100, 1, 100]) \\ \hline
Quaternion tracking cost ($Q_{q}$)  & 10000 \\ \hline
Quaternion terminal cost weight & 10000 ($Q_{t,q}$) \\ \hline
\end{tabular}
\end{table*}

\begin{figure}
\centering
\includegraphics[width=0.9\linewidth]{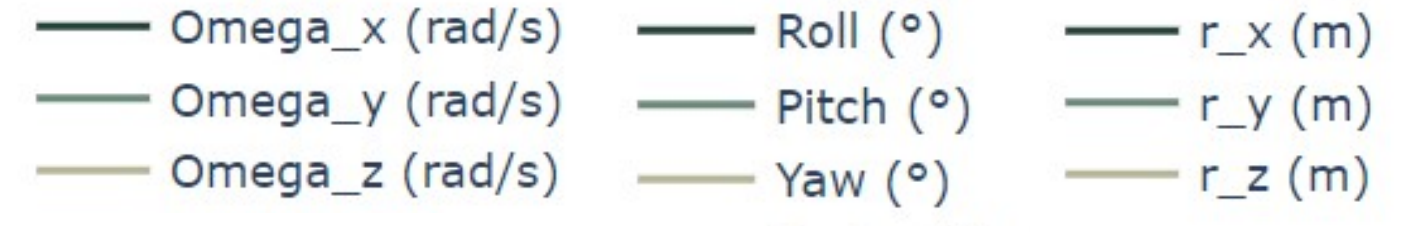}
\caption{Legend for The inertial morphing MPC results in \fig{quat_following_open}, \fig{quat_following_vel}, \fig{quat_following}.}
\label{quat_following_legend}
\end{figure}


\subsection{Launch, Inertial Morphing MPC}
\begin{figure}
\centering
\includegraphics[width=0.99\linewidth]{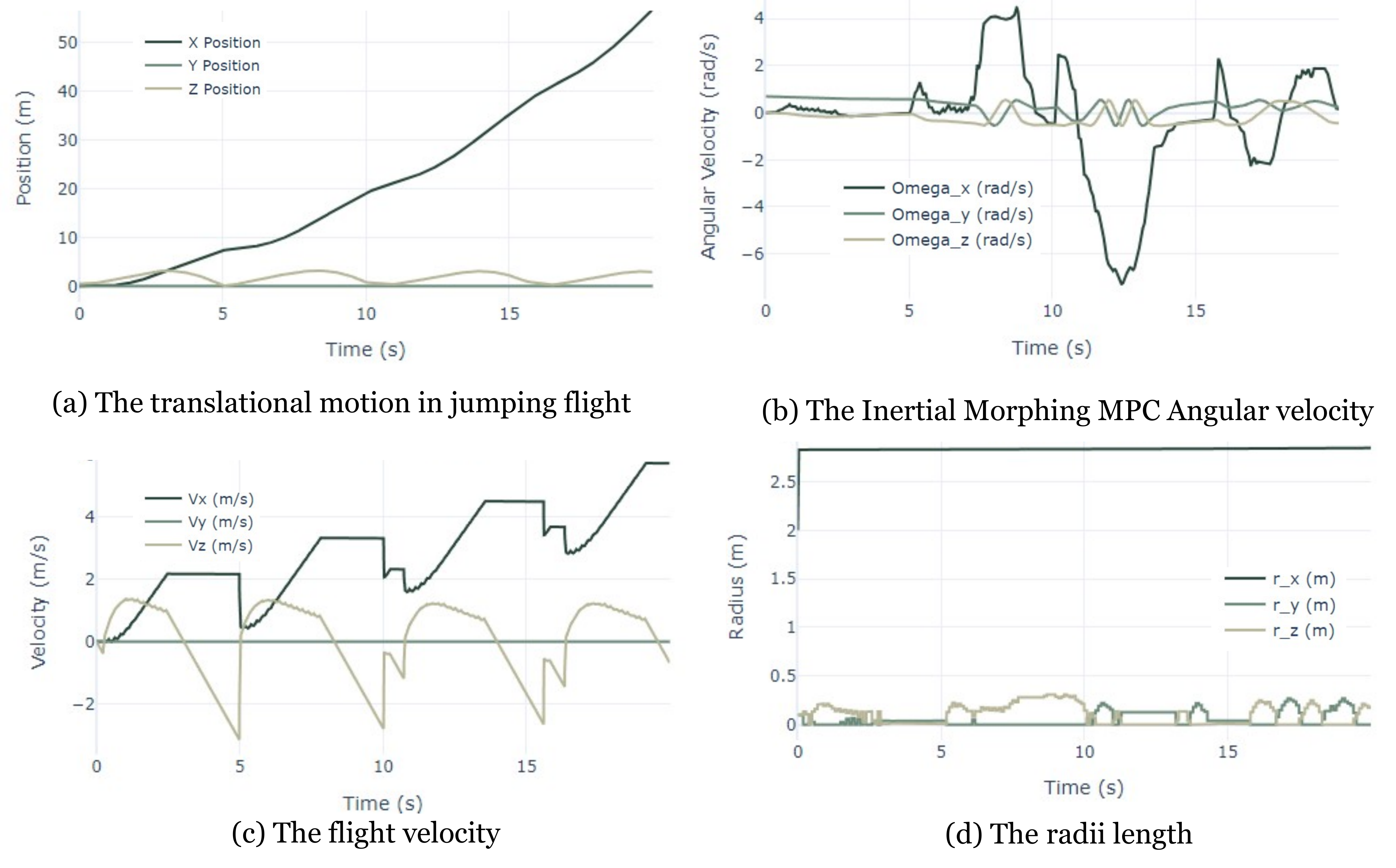}
\caption{Launch and successive jumping with the inertial morphing MPC attitude control.}
\label{full}
\end{figure}

\begin{figure}
\centering
\includegraphics[width=0.7\linewidth]{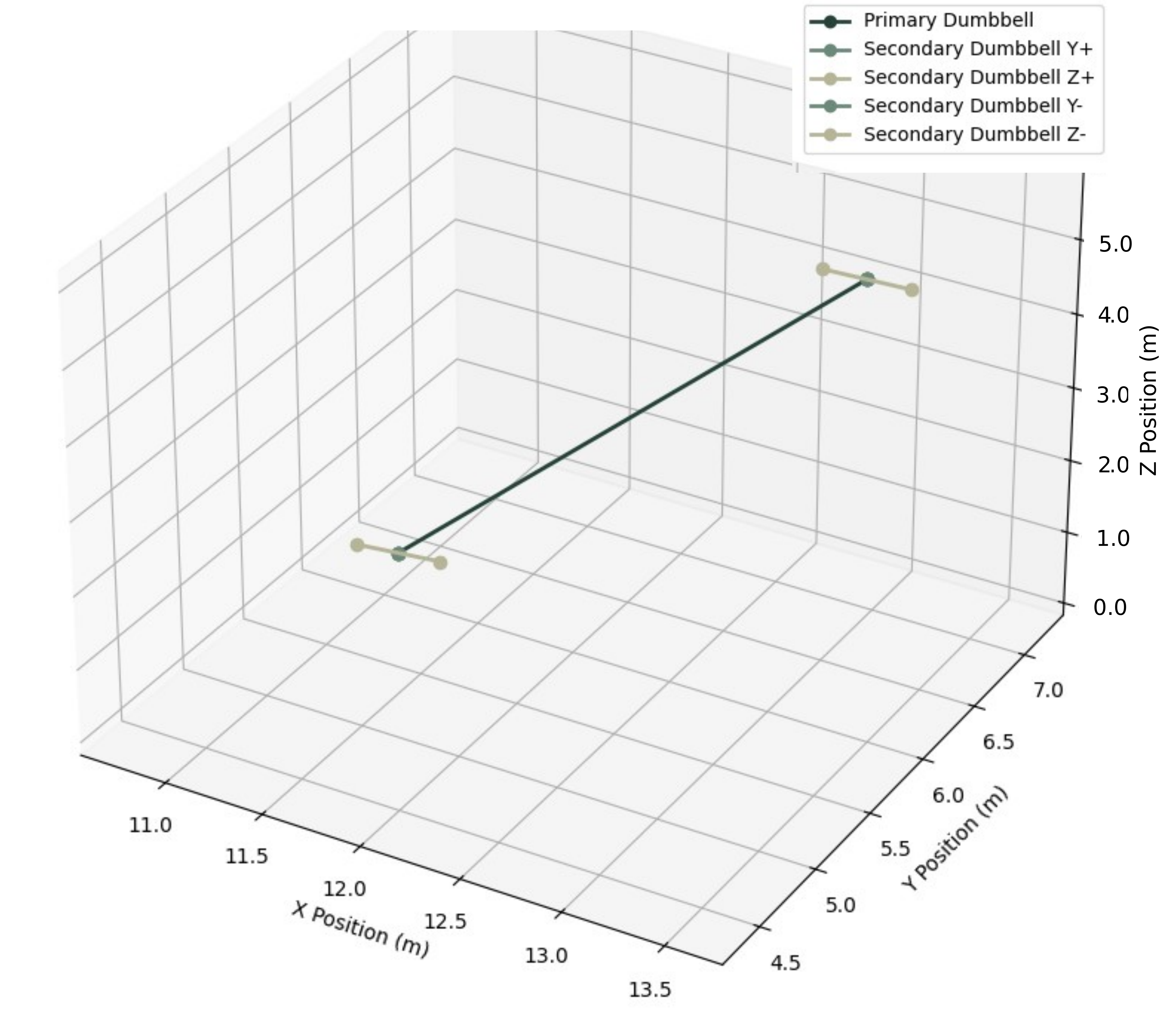}
\caption{5-Dumbbell-like \name representation jumping 3D visualization from \fig{full} at $t=8$ s. One set of legs is relatively short in length and not visible here.}
\label{dumbbell}
\end{figure}

Here, we simulate the entire sequence of the jumping flight gait under the Moon's gravity. \fig{full} shows \name successively jumps and spins and \fig{dumbbell} shows the snippet of the 3d dumbbell state. \name is moving in the $x$ direction in \fig{full}a, and the angular velocity is regulated by the inertial morphing MPC in \fig{full}b. The gravity was set to $\frac{1}{6}$ G.  

Every landing generates torque on the dumbbell-like system, causing disturbances to the angular stability. The inertial morphing MPC has stabilized sufficiently for the successive jumping operation since the MPC does not explicitly consider the contact dynamics. The dumbbell jumps at $t=0$ s and $t=5$ s as seen in \fig{full}c. The peak $v_x$ increases in the second jump, indicating that the energy from the first jump is conserved, and the second jump adds additional energy. The jumping force was set to $40$ N in the vertical and horizontal directions. \fig{full}d confirms that the range of motion for the \name is within a reasonably small range.
\name was able to achieve $2.83$ m/s horizontal jumping flight in our simulation.

\subsection{Actuator Selections}
\name limb is designed to perform both ground locomotion and dynamic jumping locomotion. On Earth, direct and quasi-direct-driven legged robots have presented agility with gear ratios between 1-10 \cite{alphred}. Under low gravity, the torque requirement is less, given the same system mass. The legged robot typically runs off a rechargeable battery, and the actuator and power circuits should be able to handle regenerative current back to the battery to store the locomotion and jumping energy. 
Considering these requirements and the calculated parameters given in Section \ref{actuator_considerations}, we chose to utilize the TBM2G-07613A stator and rotor pair for our actuators.

\subsubsection{Continuous Torque for Locomotion}

The TBM2G-07613A actuator offers a continuous stall torque of $1.23$ Nm, and $12.3$ Nm when integrated with a 10:1 gearbox, which is more than adequate for ground locomotion. Given that the robot operates in low-gravity environments, the torque required to overcome gravity is significantly lower than that required on Earth. The continuous torque rating of this actuator provides ample margin for reliable operation, ensuring that the robot can sustain locomotion without overloading the actuators.

\subsubsection{Peak Torque and Angular Velocity for Dynamic Jumping}

The actuator’s peak torque capability is critical for dynamic motions such as jumping. The TBM2G-07613A actuator has a peak stall torque of $3.43$ Nm, and combined with a gear ratio of 10:1, comfortably exceeds the calculated torque required for liftoff. The peak speed of $8000$ RPM, reduced to $800$ RPM, is still well above the calculated angular velocity of $270.3$ RPM needed for lift-off. Additionally, the regenerative current handling capabilities of the low gear-ratio BLDC actuator allows energy from the landing impact to be fed back into the robot’s battery, improving overall efficiency and energy management during prolonged operation. Dynamic and agile motion necessary for \name may concern the durability and integrity of this low gear ratio gearbox, which can be address using cycloidal gears instead of more conventional planetary or spur gears while increasing torque density \cite{zhu2024cycloidal}.

\subsubsection{Weight and Payload}
Weighing only $0.321$ kg, the TBM2G-07613A is lightweight, which is important for minimizing the overall mass of the limbed robot. Based on the continuous torque and the limb length, the maximum payload for a single quadraped is $142$kg for the static case. For the dynamic jumping case, based on the peak torque and the limb length, the maximum payload for a single quadraped is $20$kg.

\section{Discussion and Limitation}
In our inertial morphing MPC formulation, we use the principal axis inertia elements as control variables. This control approach tends to be more conservative compared to directly handling the $r_i$, though these control variables are bounded by physical inertia control mechanisms such as tether and leg lengths. This is because the MPC assumes the rate of change for the principle axis is achievable with the tether and limbs' actuator speed. Notably, the limb end-effector velocity is subject to the leg kinematics and Jacobian, introducing significant nonlinearity. To fully utilize the performance of the tether and actuator joints, the MPC would need to consider them as bounded control variables explicitly. Asymmetric inertial morphing design can reduce \name design complexity. If two bipedal robots are connected via tether instead of two quadrupeds, the number of DoF and the mass will be halved. Another potential benefit is that it can provide redundancy in attitude control for cases when one of the limbs is damaged or malfunctions.  
 A whole-body inertial morphing control is potentially possible through iteratively and locally linearizing the full dynamics as demonstrated in \cite{nmpc_contact}, though they need to be extended to handle the 3D acrobatic rotation nature of inertial morphing with contacts. Nonetheless, our simulation results indicate that the current formulation can stabilize and continue the jumping flight. We have not shown landing explicitly, which would be a part of a higher-level planning to determine the desired thrusting forces to de-accelerate the system. 

\name\ exhibits the dynamic and agile capabilities characteristic of direct and quasi-direct-driven limbed robots \cite{alphred}. Various legged and limbed robots have been proposed for planetary exploration \cite{cat_like_jump}, \cite{spacebok}, \cite{quad_jump_mpc} to address the shortcomings of existing rovers, but their robustness and reliability in actual space environments are to be tested in the future. Our simulation and controller's ability showed that the jumping flight gait with inertial morphing is viable for small, sub-$10$ kg robots. This indicates that the mini-rover multi-agent manner applies to \name, limiting the overall mission risk. Most importantly, \name's jumping flying gait offers a significantly faster means of transportation without relying on jet propulsion or aerodynamics, making it an appealing option as launch costs continue to decrease.

\section{Conclusion}
This paper explored the feasibility of successive jumping flight locomotion and inertial morphing attitude control for tethered limbed robots in low-gravity environments. Utilizing a dual-robot system connected by a tether, we demonstrated that attitude control can be potentially achieved through limb adjustments and tether length modulation without requiring traditional attitude control systems such as reaction wheels.

Simulation results indicate that the proposed system can sustain successive jumping while maintaining control over angular velocity and orientation, providing a fast and efficient method of travel for small, sub-$10$ kg planetary robots. This method shows promise for future space exploration missions where traditional rovers and aerial vehicles may not be efficient or applicable. Future work will explore experimental validation and further refinement of the MPC to improve control performance in the presence of contact dynamics and other real-world constraints.


\bibliographystyle{IEEEtran}
\bibliography{main}
\thebiography
\begin{biographywithpic}
{Yusuke Tanaka}{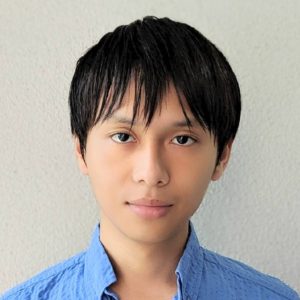} received his B.S. from UMass Amherst and is currently pursuing a Ph.D degree at the University of California, Los Angeles. He is a research assistant at RoMeLa (Robotics \& Mechanisms Laboratory) at UCLA with a research interest in contact-rich simultaneous locomotion and grasping, and limbed climbing robotics. 
\end{biographywithpic} 

\begin{biographywithpic}
{Alvin Zhu}{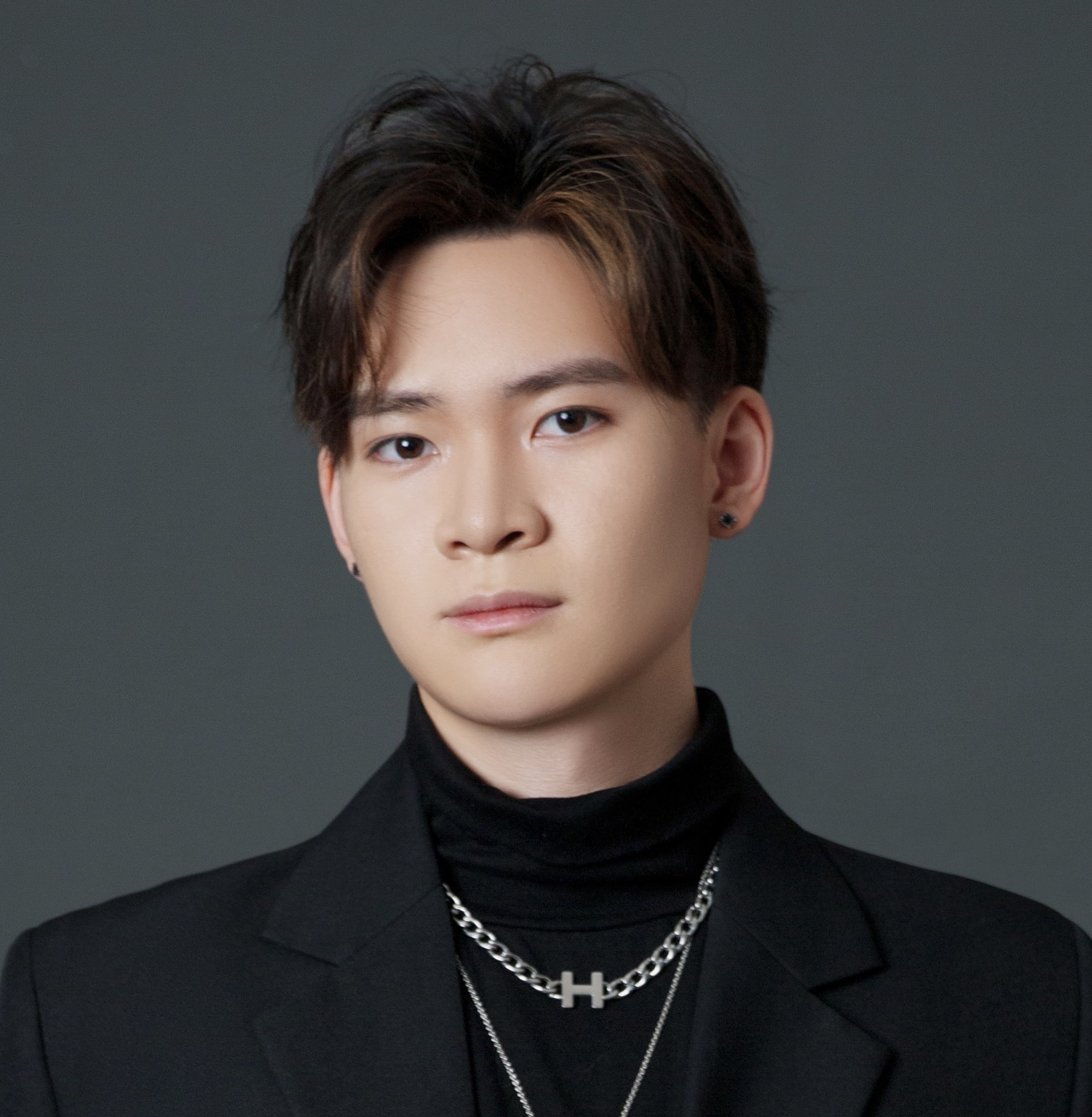}
is a member of RoMeLa (Robotics \& Mechanisms Laboratory) at the University of California, Los Angeles as a research assistant and in the Computer Science and Electrical Engineering Departments, with research interests in computer vision, learning-based controls, and reinforcement learning for humanoid robots.
\end{biographywithpic}

\begin{biographywithpic}
{Dennis Hong}{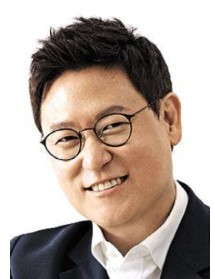} is the Founding Director of RoMeLa (Robotics \& Mechanisms Laboratory) of the Mechanical \& Aerospace Engineering Department at UCLA.
He received his B.S. degree in Mechanical Engineering from the University of Wisconsin-Madison (1994), his M.S. and Ph.D. degrees in Mechanical Engineering from Purdue University (1999, 2002). 
\end{biographywithpic}

\end{document}